# PaiP: An Operational Aware Interactive Planner for Unknown Cabinet Environments

Chengjin Wang, Zheng Yan, Yanmin Zhou*, Runjie Shen, Zhipeng Wang, Bin Cheng, Bin He.

*Abstract*—Box/cabinet scenarios with stacked objects pose significant challenges for robotic motion due to visual occlusions and constrained free space. Traditional collision-free trajectory planning methods often fail when no collision-free paths exist, and may even lead to catastrophic collisions caused by invisible objects. To overcome these challenges, we propose an operational aware interactive motion planner (PaiP) a real-time closed-loop planning framework utilizing multimodal tactile perception. This framework autonomously infers object interaction features by perceiving motion effects at interaction interfaces. These interaction features are incorporated into grid maps to generate operational cost maps. Building upon this representation, we extend sampling-based planning methods to interactive planning by optimizing both path cost and operational cost. Experimental results demonstrate that PaiP achieves robust motion in narrow spaces. Project page: https://travelers-lab.github.io/PaiP/

## I. Introduction

Box/cabinet scenarios are prevalent in human living environments due to their high space utilization efficiency and human-adapted storage organization (Fig. 1). Robots performing manipulation tasks in such scenarios face significant motion challenges, including occlusions and collisions caused by stacked objects [1, 2]. Under conditions of high visual uncertainty, a motion planner that demonstrates robust adaptability to constrained free-motion spaces is of significant importance for the deployment of robots in daily applications[3].

The key to enhancing robotic motion performance in such scenarios lies in the ability to perceive object interaction features akin to humans. Beyond visual perception, humans utilize cutaneous tactile feedback to detect force-displacement signals generated at the interaction interface due to bodily movements [4]. These signals are subsequently processed by the sensorimotor system to infer the interaction properties of objects [5-7]. Humans then leverage these interaction features to reconfigure the spatial state of objects, thereby overcoming challenges posed by unsolvable motion trajectory. Consequently, developing a tactile-based closed-loop planner offers a unique perspective for improving robotic motion adaptability in unknown and narrow environments [8]. This closed-loop sensorimotor integration enables autonomous inference of environmental interaction features through kinesthetic perception—where the robot observes the spatial state response of objects elicited by interactive stimuli to deduce their mechanical properties [9, 10]. The planner can incorporate these features as motion constraints and extrapolate them to generate motion commands. Motion directives constrained by interaction features can effectively prevent operational failures caused by unintended actions.

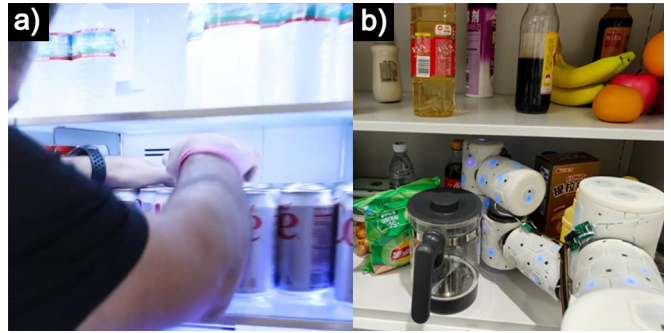

Fig. 1. Comparative motion strategies in cabinet-access scenarios for humans and robots. a) Human retrieving a beverage from a refrigerator. b) Robotic touching of a beverage from a cabinet.

The core challenge in building a tactile-based planning system lies in the appropriate representation of object interaction features. High-precision object motion models typically rely on specialized identification equipment and extensive interactive exploration, which imposes significant burdens on robots in terms of costly perceptual hardware and time consumption [11, 12]. Furthermore, this approach introduces computational challenges when the planner extrapolates these interaction features [13]. Real-time performance constitutes a critical metric for planners operating in unknown and narrow environments. Thus, balancing representation accuracy with feature complexity remains a fundamental problem in interactive motion planning.

To address this challenge, we introduce a constitutive model as the representation method for object interaction features [14]. Since the robot-object interaction operates as a low-velocity motion model with planar support, the object's motion characteristics are modeled as a spring-damper-friction system. The spring and damper components reflect the material properties of the object, while friction characterizes the interaction between the object and its environment (including the robot). This representation offers the advantage of comprehensively capturing the object's

Chengjin Wang and Zheng Yan are with the Shanghai Research Institute for Intelligent Autonomous Systems Shanghai 201210, China, with the State Key Laboratory of Autonomous Intelligent Unmanned Systems Shanghai 201210, China, and with the Frontiers Science Center for Intelligent Autonomous Systems, Shanghai 201210, China.

Yanmin Zhou, Runjie Shen, and Bin He are with the College of Electronics and Information Engineering, Tongji University, Shanghai 201804, China, with the state Key Laboratory of Autonomous Intelligent Unmanned Systems, Shanghai 201210, China, and with Frontiers Science Center for Intelligent Autonomous Systems, Shanghai 201210, China.

mechanical behavior (both motion and deformation) during interaction using a minimal set of independent parameters.

Building upon this foundation, we propose an o**p**erational **a**ware **i**nteractive motion **p**lanning framework (PaiP), as show in Fig. 2. Specifically, we develop a perception-action coordination mechanism based on motion intention, which couples kinesthetic perception with motion planning to achieve closed-loop sensorimotor integration (Fig. 2. environmental interaction feature extraction). We incorporate interaction features into the planning layer through an operational cost formulation. Consequently, we formulate the robot's motion planning problem as an optimization task that jointly considers operational cost and path cost.

Given the efficiency of sampling-based methods in high-dimensional spaces, we extend sampled planning techniques to develop an interactive motion planner (Fig. 2. interactive motion planning) [15, 16]. Unlike binary maps used for binary objects constraints, we employ a continuous grid map representation with operational cost values in the range [0,1], where 1 indicates inoperable regions and values in [0,1) represent operable areas with varying cost levels (Fig. 2. operational cost map generation). The planner's objective is to identify sampling points that optimize the operational cost. These points are then connected to form feasible paths through an optimizer, and the resulting trajectory is executed using impedance control to achieve compliant tracking (Fig.2. compliant control).

Overall, the main contributions of this work lie in: 1) a perception-action closed-loop planning framework that enhances robotic motion adaptability in constrained environments through tactile perception; 2) a multimodal tactile interaction perception method that enables autonomous extraction and real-time extrapolation of object interaction features during motion, thereby overcoming high visual uncertainty; 3) a map-based representation method for environmental interaction features, which incorporates interaction features into an operational cost map, thereby enabling sampling-based planners to generate interaction-aware motion trajectories.

## II. PROBLEM STATE AND NOTATION

This work aims to develop an interactive motion planning framework to enhance robotic motion adaptability in box/cabinet environments. Unlike boundary-constrained motion planning problems that focus on generating collision-free trajectories (geometric motion planning), interactive motion planning seeks to identify and reconfigure the spatial states of movable objects in narrow workspaces through sensory feedback. This approach overcomes challenges such as visual occlusions and unsolvable free-motion space by improving robotic adaptability in such constrained environments. Consequently, we formulate the problem as follows: given start and target states, the planning framework must simultaneously perceive and infer environmental interaction features under partially observable conditions to identify a feasible interconnected path with interactive features.

The robotic workspace can thus be represented as $\mathcal{W} \in \mathbb{R}^3$, an environment containing densely distributed objects $\mathcal{O} \in \mathbb{R}^3$ with diverse interaction features $\theta$. The manipulable regions within this environment are denoted as $\mathcal{X}_o$. The interactive motion planning problem is formally defined as a tuple $(\mathcal{X}_o, x_I, \mathcal{X}_g)$, where given a start configuration $x_I \subseteq \mathcal{X}_o$ and a target region $\mathcal{X}_g \subseteq \mathcal{X}_o$, the objective is to identify intermediate states $x_p$ within $\mathcal{X}_p \subseteq \mathcal{X}_o$ that maximize connectivity probability.

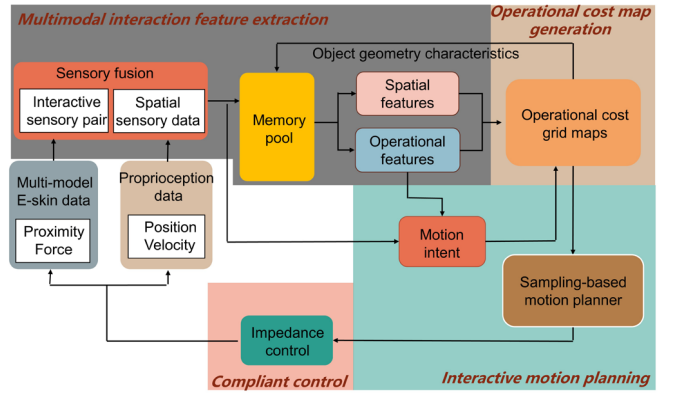

Fig. 2. Overview of the PaiP motion planning framework. The architecture comprises four components forming a closed-loop real-time planning system: multimodal interaction feature extraction, operational cost map generation, sampling-based interactive motion planning, and compliant control.

## III. RELATED WORKS

### A. Manipulation in Narrow Space

Differ from collision-free trajectory planning, robots operating in constrained spaces must accommodate controlled collisions and even leverage physical interactions to overcome the challenge of unsolvable collision-free trajectories. In recent years, research on robotic motion adaptability in cluttered environments has primarily been approached from both control and planning perspectives [13, 17-19].

To mitigate structural damage caused by catastrophic collisions in narrow spaces, compliance control methods (e.g., impedance/admittance control) have been introduced to prevent excessive contact forces during robot-object interactions [20, 21]. To address the local minima problem arising from object blockage near target points in passive compliance strategies, active compliance methods utilizing force feedback have been developed [17, 22]. Furthermore, leveraging the full-body coverage capability of electronic skins, robots have achieved whole-body compliance control[1]. The effective implementation of these control methods critically depends on the assumption of trajectory feasibility. However, in unknown environments, planners still face challenges in generating interconnected paths with interactive features based on limited local information.

To address the challenge of robotic target accessibility in narrow spaces, several studies have proposed motion planners with active interaction capabilities [13, 23]. To generate predictable interactive motion, simulation-based motion planning methods [13, 24] and model predictive planning methods [19] have been developed. Through extrapolation of the motion model, these methods can generate adaptive interactive motions tailored to the modeled object in accordance with the planned actions. However, significant challenges remain in enabling robots to autonomously construct simulation environments that meet simulator requirements or establish high-fidelity contact physics models.

*B. Sampling-based Motion Planning Methods*

Sampling-based motion planning methods have become a fundamental paradigm for robotic motion generation due to their efficiency in high-dimensional complex systems [25]. The planning objective is to generate a sequence of states through sampling that can be interconnected using a local planner. Several methods have been developed to extend motion planning to specialized scenarios, including: probabilistic roadmap planner (PRM) [26], expansive-space trees (EST) [27], rapidly-exploring random tree (RRT) [28], and batch-informed trees (BIT*) [29].

These methods utilize binary occupancy maps as environmental representations, treating environmental boundaries as constraints to generate collision-free motion commands. Recently, such approaches have been extended to narrow passages [30, 31] through the development of specialized sampling strategies, including: object-based sampling, Gaussian sampling, and bridge test sampling. Furthermore, learning-based methods have been proposed that acquire sampling distributions from demonstrations of existing sampling techniques to achieve biased sampling [32, 33]. The primary advantage of these approaches lies in their ability to achieve asymptotic optimality with reduced computational time requirements.

Ideally, incorporating environmental interaction features into the sampling map and designing constraint equations for safe interaction represents a promising approach to extend sampling-based methods from collision-free planning to efficient interactive motion planning.

IV. METHODOLOGY

In this study, a closed-loop interactive planning framework PaiP is proposed to improve the motion adaptation in narrow spaces. The method of autonomous perception of mechanical properties is described in Sec. VI-A. A network that takes proximity signals as input to predict the complete workspace topology is trained in Sec. VI-B. Subsequently, we present a method for fusing geometric and mechanical features to generate an operational cost map (Sec. VI-C). Finally, the interactive motion planning problem is formulated as a sampling-based planning task constrained by the operational cost map in Sec. VI-D.

*A. Mechanical Properties Perception*

The interaction features of objects encompass both geometric and mechanical properties. Geometric properties significantly influence the formation of interaction interfaces, while mechanical properties dominantly affect the spatial-state response of interaction forces. Therefore, this paper establishes a multi-dimensional perception and representation framework centered on interaction features, incorporating both geometric and mechanical characteristics. This framework provides essential constraints for planners to generate motions that manipulate object spatial configurations.

To autonomously infer object interaction features, robots must possess the capability to generate and acquire interactive signals at interaction interfaces. These signals should encompass interaction displacement $\triangle x$, interaction velocity $\dot{x}$, and interaction force $F$ to meet the requirements for interaction feature estimation. We formulate these perceptual data into interaction feature pairs $\psi(t)=[\triangle x_{O_i}, \dot{x}_{O_i}, f_{O_i}]^T$, where $O_i$ denotes the interaction with the $i$ object. This information can be obtained through the fusion of tactile skin sensing and proprioception. Specifically: $\triangle \mathbf{x}_{O_i} = {}^{W}P_{O_i,t} - {}^{W}P_{O_i,t-1}$, where ${}^{W}P_{O_i}$ represents the end-effector position in the world frame and $\dot{\mathbf{x}}_{O_i} = J(\mathbf{q}_t)\dot{\mathbf{q}}_t$ denotes the contact point with object $O_i$; $\dot{\mathbf{x}}_{O_i} = J(\mathbf{q}_t)\dot{\mathbf{q}}_t$, where $J(\mathbf{q}_t)$ is the Jacobian matrix at the contact point and $\dot{\mathbf{q}}_t$ is the joint velocity vector.

We formulate the inference of object mechanical properties as a parameter identification problem. These parameters represent independent physical quantities, and thus the object system is modeled as:

$$Y(t) = H(t)\theta + V(t), \qquad (1)$$

where $Y(t)=[y(1) \ ... \ y(t)]^T \in \mathbb{R}^t$ denotes the stacked output vector, $H(t)=[\psi(1) \ ... \ \psi(t)]^T \in \mathbb{R}^{t\times 3}$ represents the stacked information matrix, and $V(t)=[v(1) \ ... \ v(t)]^T \in \mathbb{R}^t$ is the stacked noise vector following a normal distribution with zero mean and variance $\sigma^2$. We employ the least squares method to estimate the object's interaction feature vector $\theta$ by minimizing the loss function:

$$J_1(\theta) = \sum_{j=1}^{t} [y(j) - \psi^T(j)\theta]^2. \qquad (2)$$

$$\left.\frac{\partial J_1(\theta)}{\partial \theta}\right|_{\theta=\hat{\theta}} = -2H_t^T(Y_t - H_t\theta)\big|_{\theta=\hat{\theta}} = 0 \qquad (3)$$

*B. Spatial Topology Network*

To construct an object-centric spatial topology of the workspace under partially observable information, we developed a spatial topology prediction network. This network predicts object spatial configurations based on local grid maps generated from limited proximity sensory signals.

Specifically, the perceived surface position of object $O_i$ is transformed into the world coordinate frame as ${}^{W}P_{O_i} = {}^{W}T_{E_i}{}^{E_i}P_{O_i}$, where ${}^{W}T_{E_i}$ denotes the homogeneous transformation matrix of the electronic skin relative to the world frame (obtained through robot forward kinematics), and ${}^{E_i}P_{O_i}$ represents the position of object $O_i$ relative to the electronic skin unit $E_i$ (measured by proximity sensors). These spatial points are subsequently converted into grid coordinates to generate local spatial grid maps. Consequently, a grid topology network (GridTopoNet) operating under partially observable conditions is proposed. GridTopoNet predicts the grid map of the entire workspace $g_{output}$ based on partially observed binary grid maps $g_{input}$. The network architecture, consists of three primary components:

- Encoder Pathway: Three convolutional blocks with 3×3 kernels and ReLU activations, each followed by 2×2 max-pooling for progressive down sampling. Each block contains two convolutional layers with batch normalization.
- Context-Aware Bottleneck: Three dilated convolutional layers (dilation rates: 2, 3, 1) to capture multi-scale contextual information while maintaining spatial resolution.
- Decoder Pathway: Three transposed convolutional blocks with 2×2 up sampling and skip connections from corresponding encoder layers. Feature fusion occurs via channel-wise concatenation of encoder and decoder features.

The final prediction head uses a 3×3 convolution with sigmoid activation to generate per-pixel occupancy probabilities (0-1 range).

Training samples are generated autonomously using a self-supervised paradigm. The trainer creates partial observation maps form random generated complete binary maps with 90% masked rate. The partial observations map serves as the network input, while the complete ground truth map functions as the training label. The training objective is to minimize the dissimilarity between the predicted map $g_{output}$ and the labeled ground truth map $\hat{g}$. For the similarity prediction training of a binary map, we adopt the Mean Squared Error (MSE) as the loss function:

$$\mathcal{L}_{loss} = \frac{1}{n}\sum_{i=1}^{n}(y_i - \hat{y}_i)^2, \quad (4)$$

*Interactive Constrains for Motion Panning*

To incorporate multi-dimensional environmental features into the constraint equations of sampling-based methods, we propose an operational-cost grid map. Unlike binary maps that represent objects and free space with values of 1 and 0 respectively, our approach employs a continuous representation in the range [0,1] based on operational cost. Specifically, free space is assigned an operational cost of 0, while manipulable regions are normalized to values within [0,1). The operational characteristics are encoded in the map as an operational difficulty metric:

$$D = \frac{\sum \theta_{O_i}}{\sum \theta_{\max}} \subseteq [0,1], \quad (5)$$

where $\theta_{\max}$ denotes the robot's interaction capability limit (a hyperparameter). This enables motion planning algorithms to reference interaction costs rather than merely avoiding collisions based on object boundaries, thereby facilitating the generation of feasible interactive trajectories.

To address the limitation of tactile perception's low spatial resolution, we introduce a memory pool module that functions as an environmental feature storage system utilizing an object-centric approach for geometric and interaction feature representation. This object-centric storage method facilitates workspace topology construction and reduces data storage requirements. Unlike visual perception, proximity sensing possesses relatively low spatial resolution, making temporal signal accumulation an effective means to mitigate perceptual uncertainty. Temporal interaction data will be utilized to infer object interaction characteristics.

*D. Sampling-based Interactive Planner*

In unknown environments, robots require additional perceptual actions to infer object interaction features. The planner must reconcile the inherent conflict between perception-driven motion and motion-constrained perception to establish an action-guided perception and perception-guided action closed loop.

To address this, we incorporate a motion intention module into the planner. This module aims to generate target points for the planner's trajectory through a perception confidence algorithm $Cov(\theta_t, \theta_{t-1})$. It takes the memory pool as input and outputs target points for both interactive perception and motion tasks. When unmodeled objects are detected in the motion path, the module switches from motion planning to kinesthetic perception [34], setting the object's center point as the target. During interaction with objects, the algorithm continuously computes the confidence matrix $\mathbf{P}_t = Cov(\theta_t, \theta_{t-1})$ between current estimates and previous measurements. When the perception covariance falls below a predetermined threshold $Tr[\mathbf{P}_t] \leq T_{conf}$, the system switches back to the motion planning target.

The planner's objective is to find the spatial state with maximum connectivity probability under given target points and operational map constraints, moving the robot from start to target positions. Leveraging the efficiency of sampling-based methods in high-dimensional complex systems, we employ such algorithms as our planner. The resulting path is then smoothed through trajectory optimization before being input to an impedance controller to ensure compliant motion during physical interaction.

V. EXPERIMENTS

In the experimental section, we focused on validating PaiP's real-time planning performance under partially observable conditions, including its motion adaptability in dense environments, the advancement of the methodology, and its robustness in physical settings. Building upon our previously developed multimodal e-skin system [35], we established both simulation and physical environments for performance evaluation, and deployed PaiP on these two platforms. For simulation tests, we constructed a cabinet scenario in the PyBullet simulator [36], where the simulated robot was equipped with proximity and force sensing capabilities comparable to those of the physical robot.

We first conducted a set of orthogonal experiments to identify a planner that meets real-time requirements from existing sampling-based planning methods. Stress tests were employed to evaluate the motion adaptability of PaiP by randomly generating test scenarios containing numerous obstacles in the simulation environment. In baseline comparisons, we performed comparative experiments between PaiP and both model-based [37] and simulation-based [13] methods consistent with the existing methodologies summarized in the related work section III.

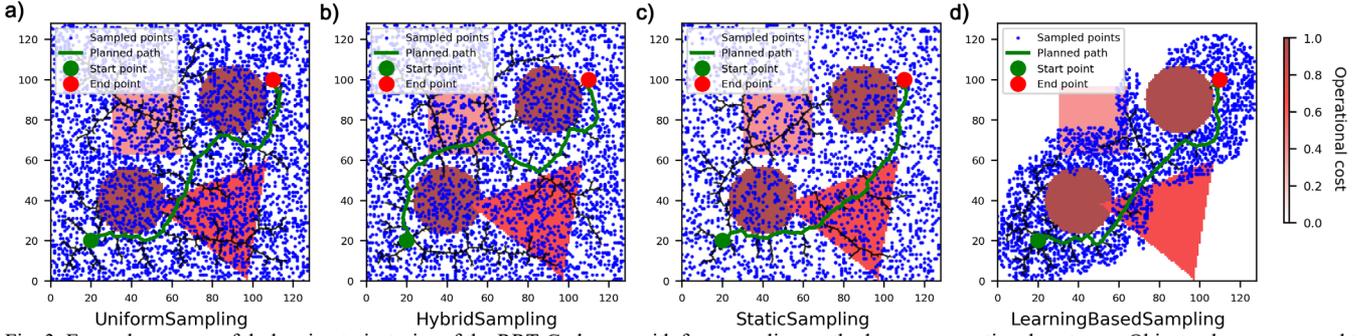

Fig. 3. Exemplary successful planning trajectories of the RRT-C planner with four sampling methods on an operational cost map. Object colors correspond to the operational cost scale at right, where 1 indicates inoperable objects. a) Sampling distribution and path generated by RRT-C with uniform sampling. b) Results obtained with hybrid sampling. c) Planning outcome using static sampling. d) Performance with learning-based sampling.

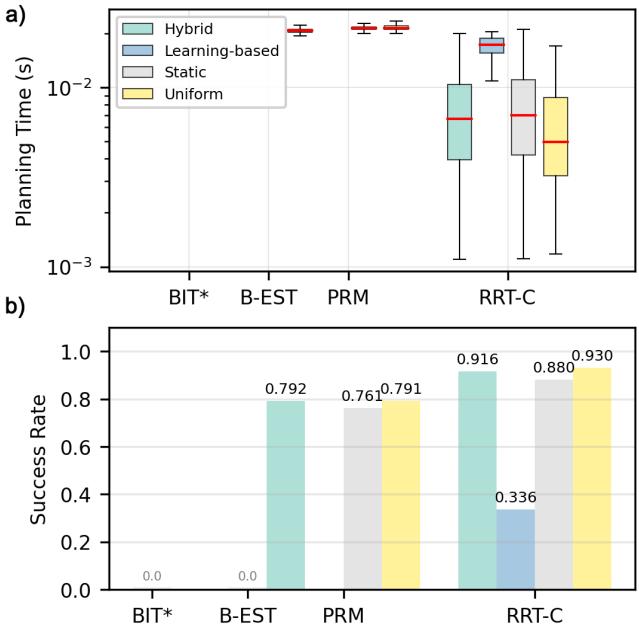

Fig. 4. Performance statistics of 16 planning-sampling combinations. (a) Computation time distribution for successful trials across all planning configurations. (b) Success rate comparison among the 16 combined planning strategies.

### A. Interactive Planner Optimization

A critical challenge in real-time planning for narrow environments is computational efficiency. Since the planner operates with only partial environmental observation, real-time updating of interaction features and subsequent replanning are essential for robots to avoid destructive actions. According to prior research, a planning frequency of 50 Hz is required to meet the stability requirements of dynamic systems [38]. Developing computationally efficient planning schemes is therefore fundamental to deployment under limited perceptual information.

To identify a planner configuration that meets real-time requirements from existing sampling-based planning and sampling methods, we developed a Python-based library integrating sampling-based motion planning algorithms with unified interfaces. This library enables flexible combination of planning methods and sampling strategies to identify optimal configurations. The repository is available at: https://github.com/Travelers-lab/sampling_based_planner_library.git. We selected four planning methods (RRT-C, B-EST, BIT*, PRM) and four sampling strategies (Uniform, Hybrid, Static, Learning-based) for combinatorial testing with total of 16 planning strategies. These methods have demonstrated exceptional performance in prior studies and represent active research directions.

We utilized randomized generation methods to create grid maps with operational costs, where objects exhibit varying operational costs and geometries. From the reachable areas of each map, we randomly selected four pairs of start and target points, ensuring a minimum separation of 50 grid units between each pair. A total of 100 maps were generated, enabling comprehensive evaluation of 16 algorithm combinations through 6,400 numerical simulations. The planning time was strictly limited to 0.02 seconds, with any exceeding duration classified as planning failure.

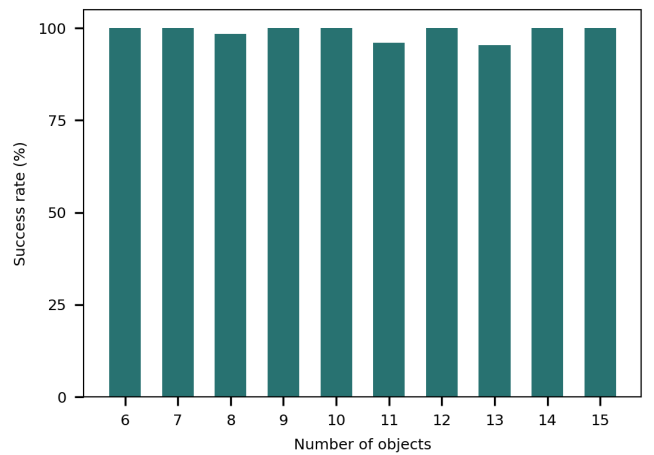

Fig. 5. Success rate statistics from PaiP stress tests.

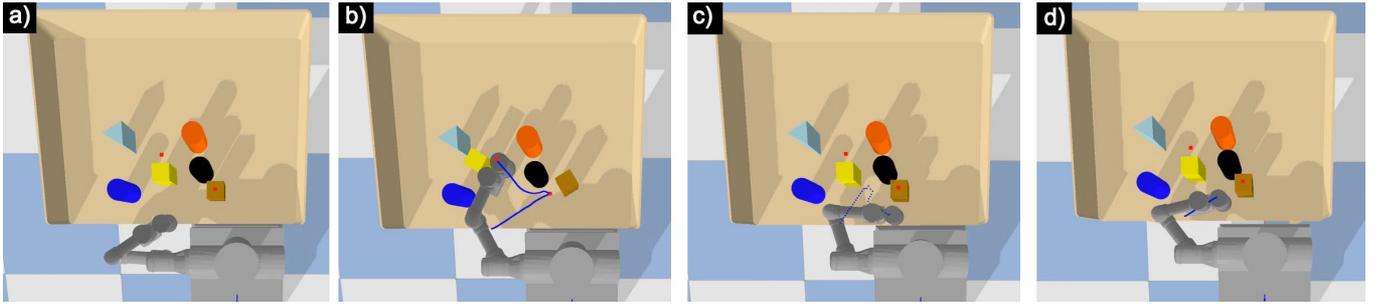

Fig. 6. Representative results from baseline comparative experiments. Red points indicate target positions for the robot, and blue trajectories represent planned paths. a) Initialized PyBullet simulation test scenario. b) Result from the PaiP planning method. c) Performance of model-based planning approach. d) Outcome of simulation-based planning method.

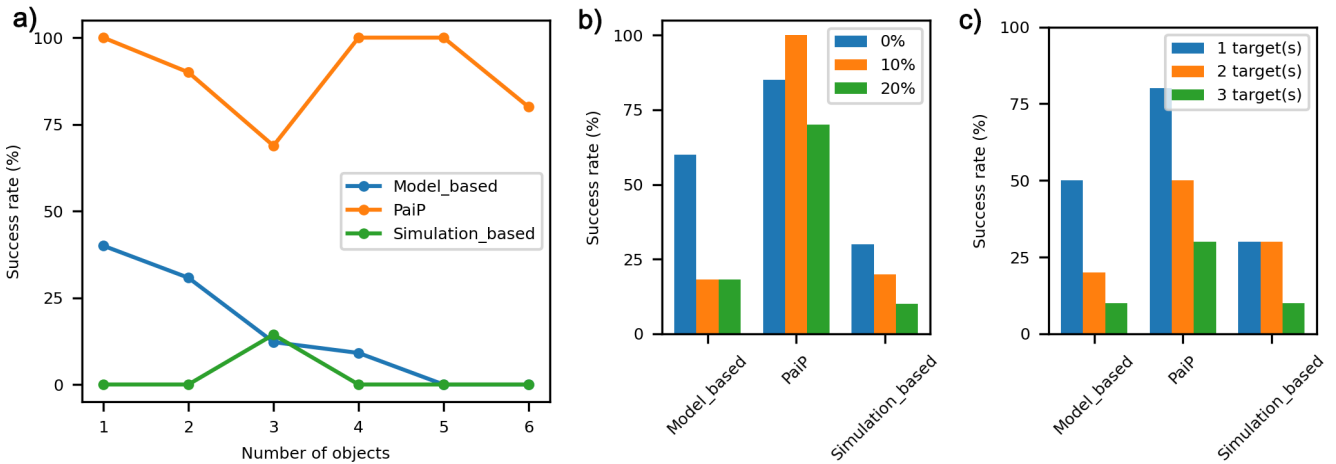

Fig. 7. Statistical results from baseline comparative experiments. Evaluated methods include PaiP, simulation-based, and model-based planning approaches. (a) Success rates under varying object numbers. (b) Performance comparison across different fixed probabilities. (c) Success rate statistics with varying numbers of target points.

Experimental results demonstrate that RRT-C, leveraging its dual-tree structure, achieved superior planning efficiency among all evaluated planners. Fig. 3 illustrates representative outcomes of RRT-C with four sampling methods. Compared to other strategies, the learning-based sampling method—by incorporating prior planning experience—produced more concentrated sampling regions, significantly reducing the search space (Fig. 3d). However, this experience-driven sampling requires additional computational overhead for model inference, resulting in inferior real-time performance (success rate: 33.6%) compared to other combinations (Fig. 4a). Overall, uniform sampling yielded optimal performance in both planning time and success rate (93%). Hybrid sampling achieved comparable success (91.6%) with fewer sampling points (Fig. 4b). RRT-C consistently outperformed three other planners across all sampling methods, while BIT* and B-EST failed to complete planning tasks within the time constraint (Fig. 4a).

Based on these numerical simulations, our PaiP adopts the combination of RRT-C with uniform sampling method

### B. Ablation Study

This ablation study is designed to identify the optimal hyperparameter configuration for GridTopoNet that can effectively capture grid map topology. The experiment focuses on two critical architectural hyperparameters: initial channels ($C_{init}$) and bottleneck channels ($C_{bot}$), which fundamentally determine the model's capacity and computational complexity.

A full factorial design encompassing all 3 combinations of initial channels {4,8,16} and bottleneck channels {16,32,64} ensures comprehensive coverage of the hyperparameter landscape. We establish a rigorous assessment framework incorporating three core performance indicators: model complexity (parameter count), inference efficiency (inference time), and prediction accuracy (MSE loss). The experimental results as shown in TABLE I:

TABLE I. ABLATION STUDY RESULTS

| $C_{init}\_C_{bot}$ | Performance Indicators | | |
|---|---|---|---|
| | Parameter Count | Inference Time (ms) | MSE Loss |
| 4-16 | 7397 | 0.253 | 0.0388 |
| 8-32 | 29321 | 0.261 | 0.0284 |
| 16-64 | 116753 | 0.269 | 0.0284 |

Based on the results of the ablation study, an initial channel number of 8 and a bottleneck channel number of 32 are determined as the architectural hyperparameters of the model.

### C. Stress Testing

Capitalizing on the simulation's capability for large test site generation, we implemented cluttered cabinet scenarios in PyBullet for stress testing and baseline evaluation. This experimental setup specifically investigates the impact of object density on PaiP's performance by varying object counts within [6, 15] at zero fixed probability. The volume of these objects ranges from 38% to 95% of the total workspace. The planner must plan interactive trajectories with objects to reach target positions.

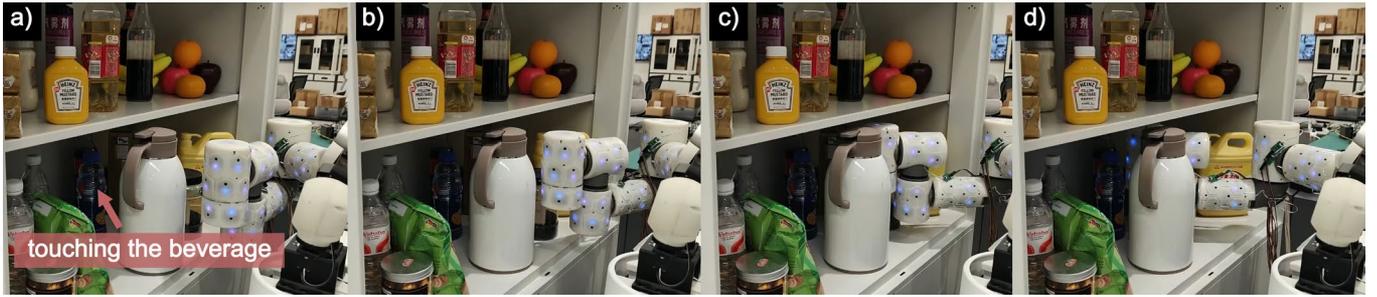

Fig. 8. Motion sequence of PaiP in physical environment testing. (a) Initial task scenario with the objective of touching a beverage inside the cabinet. (b) Planned avoidance trajectory after detecting the water-filled bottle exceeds manipulation limits. (c) Spatial reconfiguration of the plastic bottle to expand free-motion space. (d) Successful completion of the motion planning task in narrow environment.

Stress tests reveal that increasing movable object counts has negligible impact on PaiP's success rate. The framework maintains exceptional performance with success rates consistently between 95.4% and 100% across all tested configurations (6-15 objects). This robustness stems from PaiP's ability to plan and perceive the effects of interactive maneuvers, acquiring interaction features that are incorporated into the planner as operational costs, thereby enabling autonomous navigation in such narrow spaces (Fig. 5).

### D. Baseline Comparison

To benchmark the proposed method against existing planning approaches in cabinet environments and evaluate the performance boundaries of our PaiP framework, we developed a simulation testbed based on our previously established multimodal electronic skin platform (Fig. 1b).

We considered target positions, object locations, fixed probability, and object contours as variables for generating test scenarios. In baseline testing, we employed a controlled variable approach: first examining the impact on success rates with object quantities ranging [1-6], fixed ratio of 0.1, and single target configuration; subsequently evaluating fixed probability values [0, 0.1, 0.2, 0.3] with 3 objects and single target; finally testing multiple target configurations (3 targets) with 3 objects and 0.1 fixed probability.

Under all baseline comparison conditions, PaiP demonstrated significantly superior planning performance compared to baseline methods. A representative experimental result shows that when the target point was occupied by a movable object, both simulation-based and model-based planning methods failed due to inability to find feasible paths (Fig. 6), whereas PaiP successfully overcame this challenge through active physical interaction. We refined this process during our experiments.

In the object quantity baseline tests, PaiP achieved average success rates 74.5% and 87.4% higher than simulation-based and model-based methods respectively (Fig. 7a). The number of objects exhibited minimal impact on PaiP's performance, consistent with the stress test simulation results. Similar to simulation-based and model-based planning methods, PaiP showed sensitivity to two test variables: object fixation probability and number of target points (Fig. 7b, c). Success rates gradually decreased as the difficulty level of these variables increased.

### E. Physical Environments Testing

To validate PaiP's robustness in physical environments, we selected a household cabinet as the test scenario. The task required the robot to touch a beverage inside the cabinet, simulating human grasping actions. As shown in Fig. 8a, the cabinet contained densely arranged household items, eliminating feasible collision-free trajectories for task execution.

Experimental results demonstrate that upon contacting a water-filled bottle, the robot inferred the container's interaction features and recognized its manipulation limits. Consequently, PaiP generated a trajectory circumventing the bottle (Fig. 8b). When interacting with a plastic bottle, the system updated its interaction features and reconstructed its spatial configuration, thereby expanding the free-motion space (Fig. 8c). Ultimately, the robot successfully touched the target beverage bottle through compliant tracking of the interaction trajectory planned by PaiP (Fig. 8d).

## VI. CONCLUSION

This paper proposes PaiP, a multimodal tactile perception-based motion planning framework for box/cabinet scenarios. Simulation and experimental results demonstrate that the framework enables robots to effectively overcome challenges of visual occlusion and constrained free space. This approach significantly enhances the robot's motion adaptability in such environments.

In future work, considering the superior generalization capabilities of learning-based planning methods, we will attempt to utilize the planned trajectories from PaiP as motion experience to train an end-to-end model for interactive motion planning.

ACKNOWLEDGMENT

This work was supported in part by the National Key Research and Development Program of China (No. 2024YFB4709800), in part by the National Natural Science Foundation of China (No. 62473294, 62088101), in part by